\documentclass[manuscript,screen]{acmart}
\AtBeginDocument{%
  }

\setcopyright{acmlicensed}
\copyrightyear{2025}
\acmYear{2025}
\acmDOI{XXXXXXX.XXXXXXX}
\acmConference[Conference acronym ECAI-2025]{European Conference on Artificial Intelligence, CLEAR-AI Workshop: Collaborative Methods and Tools for Engineering and Evaluating Transparency in AI}{October 25--30, 2025}{Bologna, Italy}

\usepackage{listings}
\usepackage{graphicx} 
\usepackage{caption}  
\usepackage{subcaption} 
\usepackage{amsmath}
\usepackage{wrapfig}
\usepackage{hyperref}

\begin{document}

\title{Uncertainty-Guided Expert-AI Collaboration for Efficient Soil Horizon Annotation}


\author{Teodor Chiaburu}
\orcid{0009−0009−5336−2455}
\email{chiaburu.teodor@bht-berlin.de}
\affiliation{
    \institution{Berliner Hochschule für Technik} 
    \city{Berlin} 
    \country{Germany}}

\author{Vipin Singh}
\orcid{0000-0002-4472-8285}
\email{vipin.singh@bht-berlin.de}
\affiliation{
    \institution{Berliner Hochschule für Technik} 
    \city{Berlin} 
    \country{Germany}}

\author{Frank Haußer}
\orcid{0000−0002−8060−8897}
\email{frank.hausser@bht-berlin.de}
\affiliation{
    \institution{Berliner Hochschule für Technik} 
    \city{Berlin} 
    \country{Germany}}
\authornote{Equal supervision.}

\author{Felix Bießmann}
\authornotemark[1]
\orcid{0000−0002−3422−1026}
\email{felix.biessmann@bht-berlin.de}
\affiliation{
    \institution{Berliner Hochschule für Technik and Einstein Center Digital Future} 
    \city{Berlin} 
    \country{Germany}}

\renewcommand{\shortauthors}{Chiaburu et al.}

\begin{abstract}
    Uncertainty quantification is essential in human-machine collaboration, as human agents tend to adjust their decisions based on the confidence of the machine counterpart. Reliably calibrated model uncertainties, hence, enable more effective collaboration, targeted expert intervention and more responsible usage of Machine Learning (ML) systems. Conformal prediction has become a well established model-agnostic framework for uncertainty calibration of ML models, offering statistically valid confidence estimates for both regression and classification tasks. In this work, we apply conformal prediction to \textit{SoilNet}, a multimodal multitask model for describing soil profiles. We design a simulated human-in-the-loop (HIL) annotation pipeline, where a limited budget for obtaining  ground truth annotations from domain experts is available when model uncertainty is high. Our experiments show that conformalizing SoilNet leads to more efficient annotation in regression tasks and comparable performance scores in classification tasks under the same annotation budget when tested against its non-conformal counterpart. All code and experiments can be found in our repository: \href{https://github.com/calgo-lab/BGR}{https://github.com/calgo-lab/BGR}.
\end{abstract}




\maketitle

\section{Introduction}

Monitoring and characterizing soil profiles is essential for agriculture and environmental management. Soil health has a direct impact on each of us, since this is where most of our food is coming from. Accurate soil descriptions support decisions in land use planning and crop suitability. However, traditional soil analysis relies heavily on manual inspection by trained experts, making large-scale monitoring expensive and time-consuming. As a result, automating parts of the soil annotation process has become an important goal, especially in the face of the climate change effects, which highly affects soil quality \cite{desertification}.

Complete soil description is typically a multimodal problem \cite{kartiereranleitung6} that integrates visual input - such as soil profile photographs (see \autoref{fig:ex_intervals}) - with geotemporal context and chemical/morphological measurements. It also involves a structured series of interdependent tasks: segmenting the soil into meaningful horizons, describing each layer's morphological properties e.g. humus content, carbonate presence, and finally assigning a horizon class label. Each task depends on the output of the previous one, forming a pipeline where early-stage errors can propagate downstream. In such settings, uncertainty is inevitable, especially given the high complexity of the label taxonomy.

To ensure that the desired automated systems can be reliably integrated into expert workflows, it is necessary to design tailored human-AI collaboration strategies that foster transparency and trust. Central to this is the ability of the AI system to effectively communicate its uncertainty to the human stakeholders involved. When uncertainty is well-calibrated and interpretable, experts can make informed decisions about when to rely on the model’s outputs and when to intervene. This not only improves overall system performance but also encourages responsible deployment in high-stakes use cases.

In this work, we make the following contributions:
\begin{enumerate}
    \item We conformalize an existing multimodal, multitask solution for describing soil horizons - \textit{SoilNet} \cite{soilnet} - to provide calibrated uncertainty estimates for two of its three tasks: depth marker prediction and horizon classification. The model's design already mirrors the workflow of human geologists; by equipping it further with calibrated uncertainties, we aim at rendering the model's reasoning process even more transparent.
    \item We propose an evaluation protocol inspired by a real-world application of the proposed model, assuming a limited annotation budget, which enables systematic evaluation of different uncertainty quantification strategies across both regression and classification tasks. This framework allows us to study how uncertainty-aware collaboration improves annotation outcomes while minimizing expert involvement.
\end{enumerate}

\section{Related Work}

\paragraph{Active Learning.} 
Active Learning (AL) is a paradigm in Machine Learning (ML) where the model selectively queries the most informative data points for annotation, with the goal of improving performance using fewer labeled samples \cite{settles_active}. This approach is particularly valuable in domains where labeling is costly or requires expert knowledge. Classical strategies include \textit{uncertainty sampling}, where the model queries examples on which it is least confident \cite{lewis_uncert_sampling,liu_uncert_sampling} and \textit{query-by-committee}, where multiple models "vote" and samples with high disagreement are selected \cite{seung_query_committee,freund_query_committee}. Inspired by the evaluation protocols of active learning strategies we also assume a limited labeling budget. The key difference to our work is that while active learning requires a limited label budget during training, we assume such a budget at inference time to minimize the human expert annotation workload.

\paragraph{Quantifying Epistemic Uncertainty.} 
Epistemic uncertainty reflects the model's lack of knowledge and is particularly important in sensitive applications where incorrect predictions can have serious consequences. To estimate this form of uncertainty, ML systems must move beyond point predictions and instead produce a distribution over possible outputs. Approaches to estimating output distributions of ML models can be broadly categorized in model specific uncertainty estimates and model agnostic approaches. Model specific approaches often use dedicated uncertainty aware models, such as \textit{Bayesian Neural Networks} (BNNs), often approximated using methods such as Monte Carlo Dropout (MCD) \cite{mcd,PFMCD}, Deep Ensembles \cite{deep_ensembles} or Variational Inference \cite{blundell_var_infer,graves_practical_2011}. Model agnostic approaches instead can be used with arbitrary ML models. A popular type of model agnostic uncertainty estimators is \textit{Conformal Prediction} \cite{vovk_book,vovk_tutorial,angelopoulos_gentle}, which offers distribution-free, finite-sample guarantees on predictive sets or intervals. Conformal methods wrap around any base model and calibrate uncertainty post-hoc using a held-out calibration set, enabling models to produce uncertainty estimates. In this work, we adapt conformal prediction to a human-AI collaborative multi-task annotation pipeline. 

\paragraph{Multitask Uncertainty and Expert HIL Strategies.} 
Recent work has explored uncertainty-aware models that defer difficult cases to human experts, especially when multiple tasks or modalities are involved. For example, \cite{conformal_hil} demonstrate that showing conformal prediction sets significantly improves human–AI team accuracy on classification, sentiment analysis and NER (Named Entity Recognition) tasks, as opposed to simple top-$K$ prediction sets. In their experiments, they use the conformal set size as a measure for the difficulty the model had to solve the task at hand and show that it correlates with human annotation accuracy (the larger the set size, the more the human performance decreases).
A related line of work is \textit{learning to defer} (L2D), where a model is trained to output “not sure – ask an expert” via a \textit{deferral loss}. In \cite{l2d}, the authors develop a two-stage L2D framework for 'multi-task medical diagnosis' that involve both classification (predicting mortality) and regression (length of hospital stay). In their setting, if the model is sufficiently confident, its diagnosis is accepted; otherwise, the decision is deferred to a medical expert.  Similarly, other works use \textit{classification with rejection}: \cite{explain_uncertain} propose a general framework where uncertain model predictions - computed e.g. via MCD \cite{PFMCD} or Deep Ensembles \cite{deep_ensembles} - are formulated as explanations and trigger human review, until a fixed \textit{labeling budget} is exhausted.
In robotics, \cite{sirius} introduce \textit{SIRIUS-Fleet} for multi-task robot learning, which uses anomaly detection across tasks to request human intervention. If any task’s predicted outcome is highly uncertain or anomalous, the system flags it and waits for a human monitor to correct its actions. Moreover, the system adapts its thresholds via human feedback, which over time leads to fewer requests for human intervention.

\section{Data}

We conducted our experiments using a multimodal dataset provided by our partner geological institute (reference anonymized). To the best of our knowledge, this is the most comprehensive image-tabular dataset available for automatic soil classification. It includes both soil profile images and expert-annotated tabular data.

The dataset contains 3349 soil profile photographs (example in \autoref{fig:ex_intervals}), captured during field expeditions after excavating 1-meter-deep pits. Each image presents a frontal top-down view of the exposed soil wall. Each image is paired with structured annotations. These include geotemporal metadata for the whole profile and horizon-level morphological properties such as color or carbonate content. In total, the dataset comprises 13621 annotated horizons, mapped to 99 clustered horizon classes. Each soil profile contains between 2 and 8 horizons. More details about the data can be read in \cite{soilnet}.

We have split the data into training, validation, calibration and test sets in ratios of 0.6 - 0.2 - 0.1 - 0.1. We trained and fine-tuned our model on the first two sets and determined the nonconformity scores for conformal prediction on the calibration set (see \autoref{subsec:conformal}). We evaluate our automatic labeling pipeline on the test set in \autoref{sec:results}.

\section{Methods}

In this section, we briefly describe how geologists commonly annotate soil horizons. Afterwards, we outline the theoretical foundations of conformalizing regressors and classifiers and, in the end, provide details regarding our experimental design.

\subsection{Multimodal Multitask Soil Description}

Describing soil profiles in the field is a structured multimodal process that geologists perform in several interdependent steps. First, the soil profile is visually segmented into discrete horizons. Next, each horizon is described in terms of key morphological properties, including humus content, carbonate presence or color. Finally, each horizon is assigned a class label from a hierarchical taxonomy. The authors in \cite{soilnet} model this process as a sequence of three tasks: (1) predicting depth markers to segment the image into horizons, (2) predicting morphological properties for each segment, (3) classifying the horizon label. Their proposed architecture - SoilNet - jointly solves these tasks using both the image and tabular inputs. In this paper, we exemplarily run experiments on a subset of the SoilNet's task solver modules: one regression task (depth marker prediction) and one classification task (the final horizon label prediction).

\subsection{Conformalizing the Task Solvers}\label{subsec:conformal}

We applied conformal prediction to the depth and horizon label prediction tasks in our soil description pipeline. Here, we describe how we conformalized the depth marker predictor (regression) and the horizon classifier. Our goal was to obtain uncertainty estimates for both tasks such that, for a user-specified confidence level $1 - \alpha$, the model's prediction set (classification) or interval (regression) contains the true value with probability at least $1 - \alpha$. We correct for the finite sample bias of the empirical quantile by multiplying the confidence level with $\frac{n+1}{n}$, where $n$ is the size of the calibration set, as recommended in \cite{angelopoulos_gentle}.

\begin{figure}[htbp]
    \centering
    \begin{subfigure}[b]{0.45\textwidth}
        \centering
        \includegraphics[width=\linewidth]{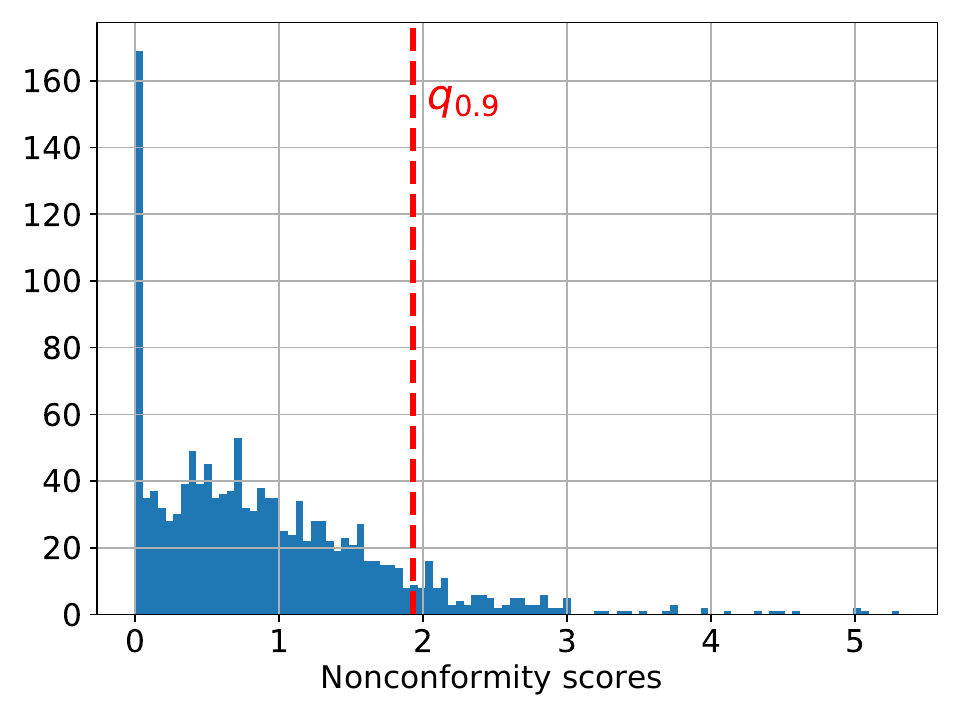}
        \caption{Depths}
    \end{subfigure}
    \begin{subfigure}[b]{0.45\textwidth}
        \centering
        \includegraphics[width=\linewidth]{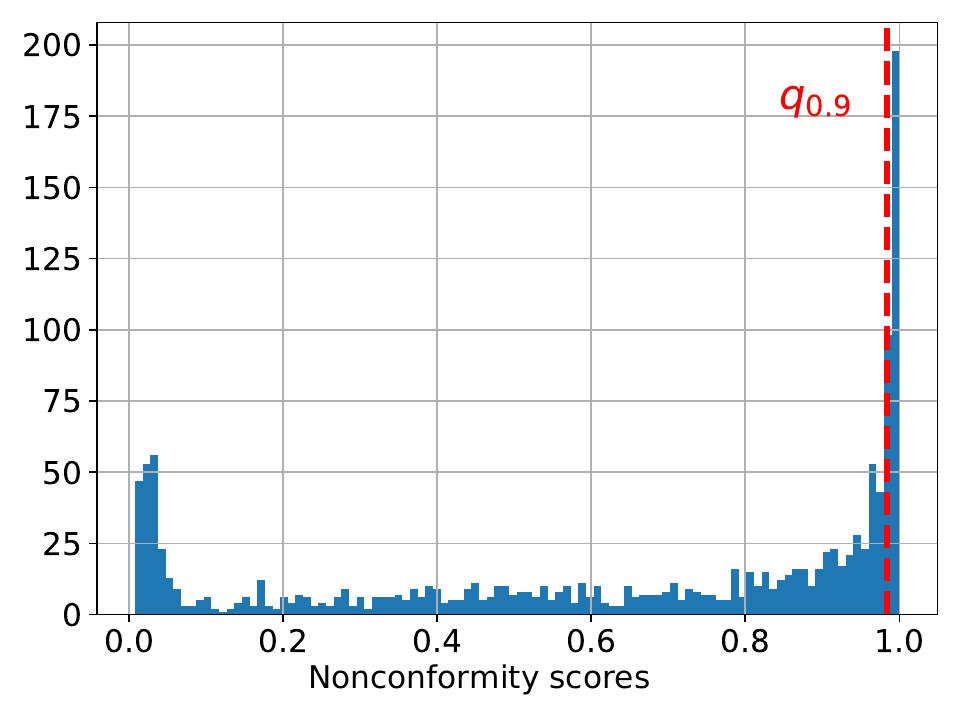}
        \caption{Horizons}
    \end{subfigure}
    \caption{Calibrating the depth marker predictor \textbf{(a)} and the horizon classifier \textbf{(b)}. In both cases, after consulting our geologist partners, we settled for a 90\% probability guarantee (so $\alpha = 0.1$). The nonconformity scores were computed as explained on the calibration set. For reference, $q_{depths} = 1.9168$ and $q_{horizons} = 0.9942$, both high quantile values being indicative of the generally high model uncertainty in both tasks.}
    \label{fig:scores}
\end{figure}

\paragraph{Depth Marker Prediction.}

\begin{figure}[htbp]
    \centering
    \includegraphics[width=\linewidth]{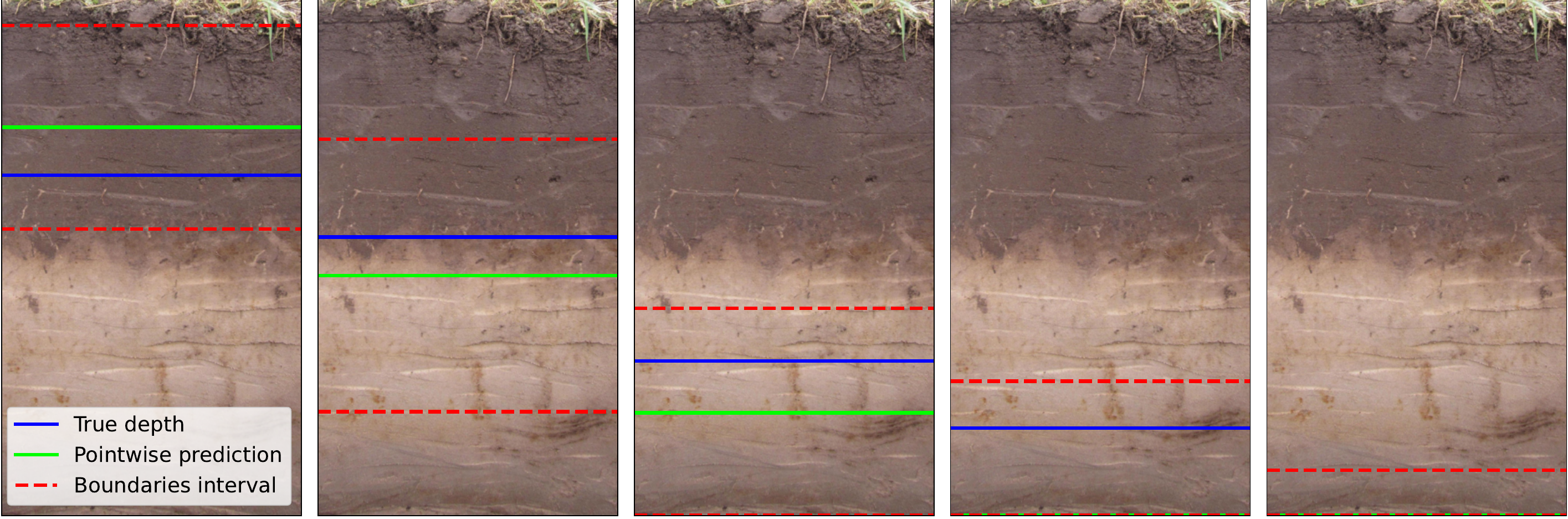}
    \caption{Soil profile example along with its five depth markers: ground truth (blue), pointwise prediction (green) and conformal interval boundaries (red). The fifth ground truth and predicted markers are at the stop token level (-1 meter). \textit{Image license anonymized.}}
    \label{fig:ex_intervals}
\end{figure}

Similar to the notation convention in \cite{soilnet}, let $\hat{d}_t^{(i)} \in (0, 1]$ denote the predicted depth marker for the $t$-th horizon in the $i$-th input $\mathbf{x}^{(i)}$ (soil sample image along with geotemporal data) and let $d_t^{(i)} \in (0, 1]$ be the corresponding ground truth marker, see \autoref{fig:ex_intervals}. Recall that the considered horizon depths only reach -1 meter, hence the interval boundaries. For the whole sample $\mathbf{x}^{(i)}$, let $\mathbf{\hat{d}}^{(i)} = [\hat{d}_1^{(i)}, \hat{d}_2^{(i)}, \dots, \hat{d}_D^{(i)}]$ denote the complete ordered list of predicted depth markers and $\mathbf{d}^{(i)} = [d_1^{(i)}, d_2^{(i)}, \dots, d_D^{(i)}]$ the actual depths\footnote{For symmetry reasons in batching, the authors in \cite{soilnet} pad the output with a stop token $s = 1$, so that every list of depth markers has a fixed length $D = 8$.}.

We use adaptive conformal regression to generate input-dependent prediction intervals by following the approach described in \cite{conf_residual}. This requires training a residual model to estimate the absolute error $|\hat{d}_t^{(i)} - d_t^{(i)}|$, using a held-out calibration set. We denote the predicted residual for the $t$-th horizon of the $i$-th input by $u_t^{(i)}$. For our experiments, we trained a 3-layer-MLP with ReLU activations in-between (more details in our repository). The inputs on which the residual model was trained consisted of the visual segment features extracted by SoilNet concatenated with the encoded geotemporal data.

Given a calibration set $\mathcal{C_d} = \{(\mathbf{x}^{(i)}, \mathbf{d}^{(i)})\}_{i=1}^n$, we compute the nonconformity scores:

\begin{equation}\label{eq:scores_dep}
    s_t^{(i)} = \frac{|\hat{d}_t^{(i)} - d_t^{(i)}|}{u_t^{(i)}}, \quad \forall i = 1, ..., n, \quad \forall t = 1, ..., D
\end{equation}

Let $S_{depths}$ be the set of all the nonconformity scores relevant for the depth predictions. Then, we determine the quantile threshold (see \autoref{fig:scores}):

\begin{equation}\label{eq:q_dep}
    q_{depths} = \text{Quantile}_{1 - \alpha}( S_{depths} )
\end{equation}

At inference time, for a new sample $\mathbf{x}^{(j)}$ from the test set $\mathcal{T_d} = \{(\mathbf{x}^{(j)}, \mathbf{d}^{(j)})\}_{j=1}^m$, we construct a conformal prediction interval for the $t$-th horizon as:

\begin{equation}\label{eq:conf_interval}
    \mathcal{I}_t^{(j)} = \left[ max(0, \hspace{0.3em} \hat{d}_t^{(j)} - q_{depths} \cdot u_t^{(j)}),\quad min(\hat{d}_t^{(j)} + q_{depths} \cdot u_t^{(j)}, \hspace{0.3em} 1) \right]
\end{equation}

This interval is guaranteed to contain the true depth marker with probability at least $1 - \alpha$ \cite{vovk_tutorial} and is adaptive to local difficulty via $u_t^{(j)}$. Given the boundary constraints of the physical problem (horizons can only be predicted within the 1-meter range), we round up or down the interval boundaries to 0 or 1, respectively, when needed. \autoref{fig:ex_intervals} shows an example of predicted conformal depth intervals.

\paragraph{Horizon Classification.}

For the horizon classification task, let $\mathbf{\hat{h}}_t^{(i)} = (\hat{h}_{t,1}^{(i)}, \hat{h}_{t,2}^{(i)}, \dots, \hat{h}_{t,H}^{(i)})$ denote SoilNet's softmax output over $H$ horizon classes for the $t$-th horizon in the $i$-th sample of the calibration set. Here, the calibration set is defined as $\mathcal{C_h} = \{(\mathbf{x}^{(i)}, \mathbf{h}^{(i)})\}_{i=1}^n$, where $\mathbf{h}^{(i)} = [h_1^{(i)}, h_2^{(i)}, \dots, h_D^{(i)}]$ is the list of ordered true labels (top to bottom) for all $D$ horizons in the $i$-th sample, with $h_t^{(i)} \in \{1, 2, \dots, H\}, \forall t$. We further define a nonconformity score for each calibration sample as the error of the model score for the true class $c := h_t^{(i)}$:

\begin{equation}\label{eq:scores_hor}
    s_t^{(i)} = 1 - \hat{h}_{t,c}^{(i)}, \quad \forall i = 1, ..., n, \quad \forall t = 1, ..., D
\end{equation}

Let $S_{horizons}$ be the set of all the nonconformity scores relevant for the horizon classification. Then, we determine the quantile threshold (see \autoref{fig:scores}):

\begin{equation}\label{eq:q_hor}
    q_{horizons} = \text{Quantile}_{1 - \alpha}( S_{horizons} )
\end{equation}

At test time, for a new sample $\mathbf{x}^{(j)}$ from the test set $\mathcal{T_h} = \{(\mathbf{x}^{(j)}, \mathbf{h}^{(j)})\}_{j=1}^m$, we construct a conformal prediction set for the $t$-th horizon by including all labels whose predicted probabilities reach at least the threshold $1 - q_{horizons}$:

\begin{equation}\label{eq:conf_set}
    \Gamma_t^{(j)} = \big\{ k \in \{1, \dots, H\} \mid \hat{h}_{t, k}^{(j)} \geq 1 - q_{horizons} \big\}
\end{equation}

Again, this prediction set is guaranteed to include the true horizon label with probability at least $1 - \alpha$ \cite{vovk_tutorial}.

\subsection{Experimental Design: Uncertainty-Guided Human-AI Collaboration}

To assess the impact of different uncertainty quantization proxies on human-machine collaboration, we simulate an annotation workflow in which a model is assisted by an expert reviewer. In this simulated setting, the model is presented with a batch of test samples and asked to solve both the regression task (depth marker prediction) and the classification task (horizon label prediction). For a limited subset of the most uncertain predictions, human domain experts provide ground truth annotations. This reflects a realistic expert-in-the-loop system constrained by a limited labeling budget.


We adopt a ranking-based selection strategy. For each task, we sort the predictions in descending order of uncertainty and simulate expert intervention on the top-$K$ most uncertain samples, where $K$ is determined by a fixed labeling budget. The most uncertain model predictions are then replaced with ground truth expert annotations for these samples, just like a real human expert would collaborate with an AI in case of high uncertainty. This allows us to quantify how well each uncertainty mitigation strategy supports the collaborative system in optimizing end-to-end annotation quality under expert workload constraints. The results, thus, have direct impact on real human-machine annotation applications, where uncertainty thresholds and labeling budgets would both be active constraints. Concretely, for an indefinite number of new (unlabeled) test samples, the expert annotator would be asked to revise samples marked as uncertain beyond a fixed threshold, as long as the query budget has not been exhausted yet. More on this in \autoref{sec:results}.

We evaluate three uncertainty ranking strategies per task:

\begin{itemize}
    \item[a)] \textbf{Depth Marker Prediction (Regression)}
    \begin{itemize}
        \item[1)] \textit{Conformal Prediction}: Rank by the width of the conformal confidence intervals $\mathcal{I}_t^{(j)}$ as given in \autoref{eq:conf_interval}.
        \item[2)] \textit{MCD Predictions}: Rank by the standard deviation of MCD predictions (the Dropout layer is active during inference).
        \item[3)] \textit{Random}: Rank samples randomly. 
    \end{itemize}
    \item[b)] \textbf{Horizon Classification}
    \begin{itemize}
        \item[1)] \textit{Conformal Prediction}: Rank by the size $|\Gamma_t^{(j)}|$ of the conformal label set in \autoref{eq:conf_set}.
        \item[2)] \textit{Softmax Output}: Rank by the entropy in the softmax output:
        \begin{equation}
            E_t^{(j)} = -\sum_{k}^H \hat{h}_{t,k}^{(j)} \log( \hat{h}_{t,k}^{(j)} )
        \end{equation}
        \item[3)] \textit{Random}: Rank samples randomly. 
    \end{itemize}
\end{itemize}

To compare the uncertainty ranking methods, we analyze how the performance -- measured as IoU for regression and accuracy, precision, recall for classification -- increases, as more human expert interventions are allowed. An efficient uncertainty quantization strategy is expected to successfully identify the most difficult samples for the model, in the absence of ground truth checks; correcting a more difficult annotation inevitably leads to a greater performance increase than correcting a simple one. In this work, similar to \cite{lewis_uncert_sampling,liu_uncert_sampling}, we link sample difficulty to model uncertainty. In other words, we define a sample as being 'difficult', whenever our proxies listed above predict that SoilNet will have a high degree of uncertainty when labeling them (assigning depth markers or horizon labels). 

\section{Results and Discussion}\label{sec:results}

In the following section we present and discuss our key findings in the simulated uncertainty-guided human-AI collaboration experiments.

\begin{wrapfigure}{R}{0.45\textwidth}
    \centering   
    \includegraphics[width=0.4\textwidth]{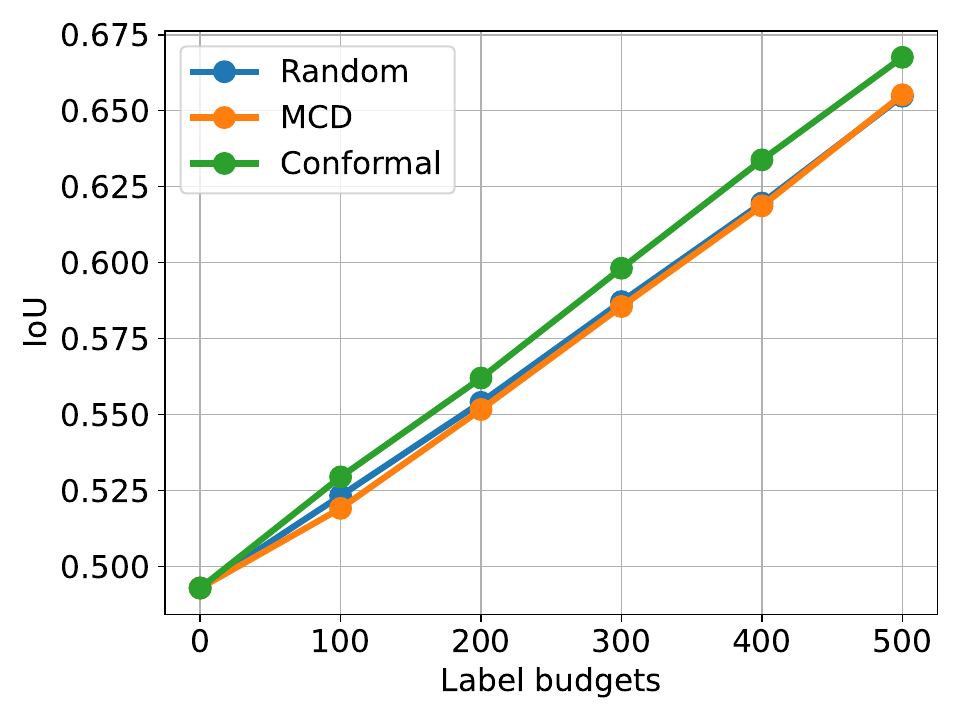}
    \caption{Correcting depth marker predictions based on conformal intervals results in a higher increase in IoU as compared to corrections based on Monte Carlo Dropout (MCD). For the random baseline, we carried out 100 simulations for each label budget and averaged the IoU. For the MCD strategy, there were 50 runs per sample. All predictions were run on the test set, which had 1364 samples in total.}
    \label{fig:task1}
\end{wrapfigure}

\paragraph{1. Regression: Conformal Intervals Estimate Uncertainty Better than MCD.} The widths of the conformal confidence intervals for regression are better at measuring uncertainty in the model's depth predictions and identifying the truly difficult samples to defer to human expert annotation. \autoref{fig:task1} reveals that ranking uncertainty w.r.t. conformal intervals is more efficient than w.r.t. the standard deviations of the MCD predictions (which are as inefficient as random queries) in terms of IoU, for all label budgets. 

\begin{figure}[htbp]
    \centering
    \includegraphics[width=\linewidth]{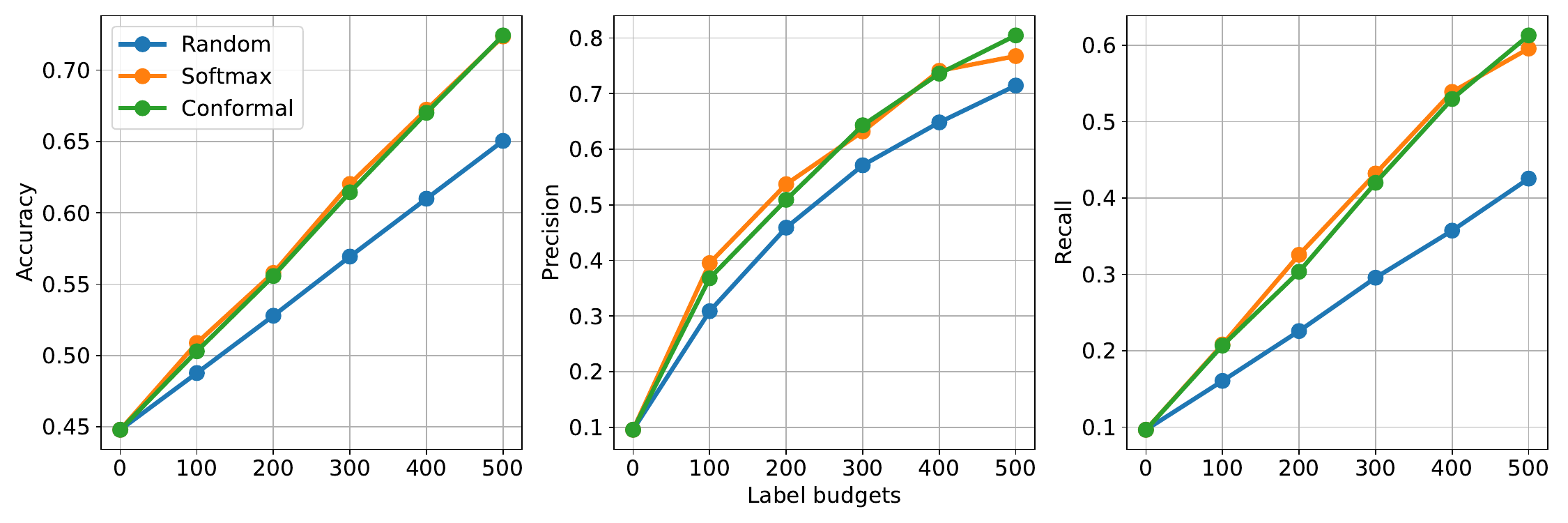}
    \caption{Correcting horizon classifications based on conformal sets and on Softmax entropies leads to similar improvements in accuracy, precision and recall. 
    All predictions were run on the test set, which had 1364 samples in total. For the random baseline, we carried out 100 simulations for each label budget and averaged the three metrics.}
    \label{fig:task3}
\end{figure}
\begin{figure}[htbp]
    \centering
    \begin{subfigure}[b]{0.45\textwidth}
        \centering
        \includegraphics[width=\linewidth]{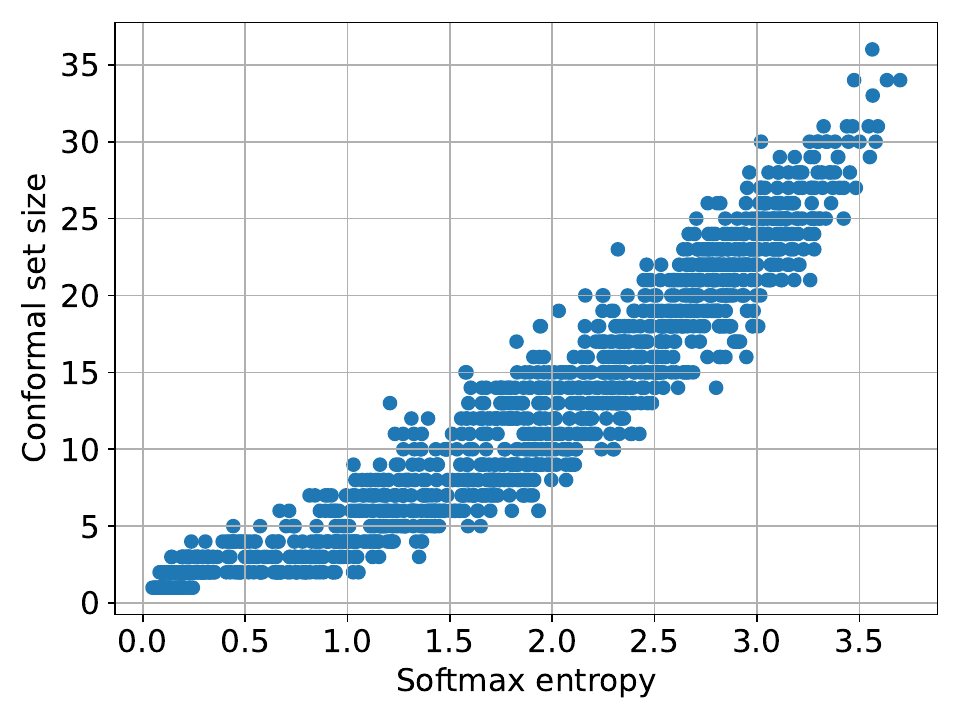}
        \caption{}
    \end{subfigure}
    \begin{subfigure}[b]{0.45\textwidth}
        \centering
        \includegraphics[width=\linewidth]{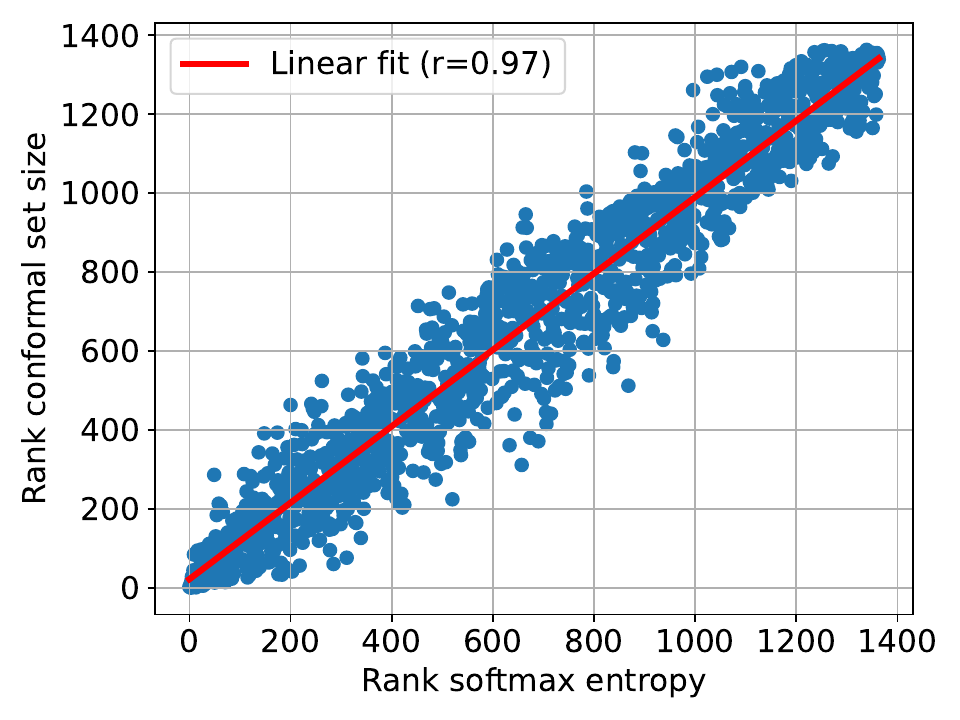}
        \caption{}
    \end{subfigure}
    \caption{Softmax entropies and label set sizes are strongly correlated. This is obvious both in \textbf{(a)}, where the two uncertainty proxies are plotted directly and reveal a non-linear dependence, as well as in \textbf{(b)}, where the ranks assigned by both methods to the test samples are highly linearly correlated. 
    }
    \label{fig:correlation}
\end{figure}

\paragraph{2. Classification: Conformal Sets and Softmax Entropies Estimate Uncertainty Equally Well.} In terms of accuracy, precision and recall, conformal label sets are as good as the Softmax entropies for measuring the model's horizon classification uncertainty, while both being better than random queries (see \autoref{fig:task3}). 
The similarities of results suggest a high correlation of the two uncertainty proxies, which proved to be, indeed, the case, as highlighted in \autoref{fig:correlation}. In \textbf{(a)}, a non-linear correlation between the Softmax entropies and the corresponding conformal set sizes is revealed. Hence, the two methods sort the query samples very similarly, which becomes even more obvious in \textbf{(b)}, where the high linear correlation of their proposed ranks is demonstrated. 

\begin{figure}[htbp]
    \centering
    \includegraphics[width=0.75\linewidth]{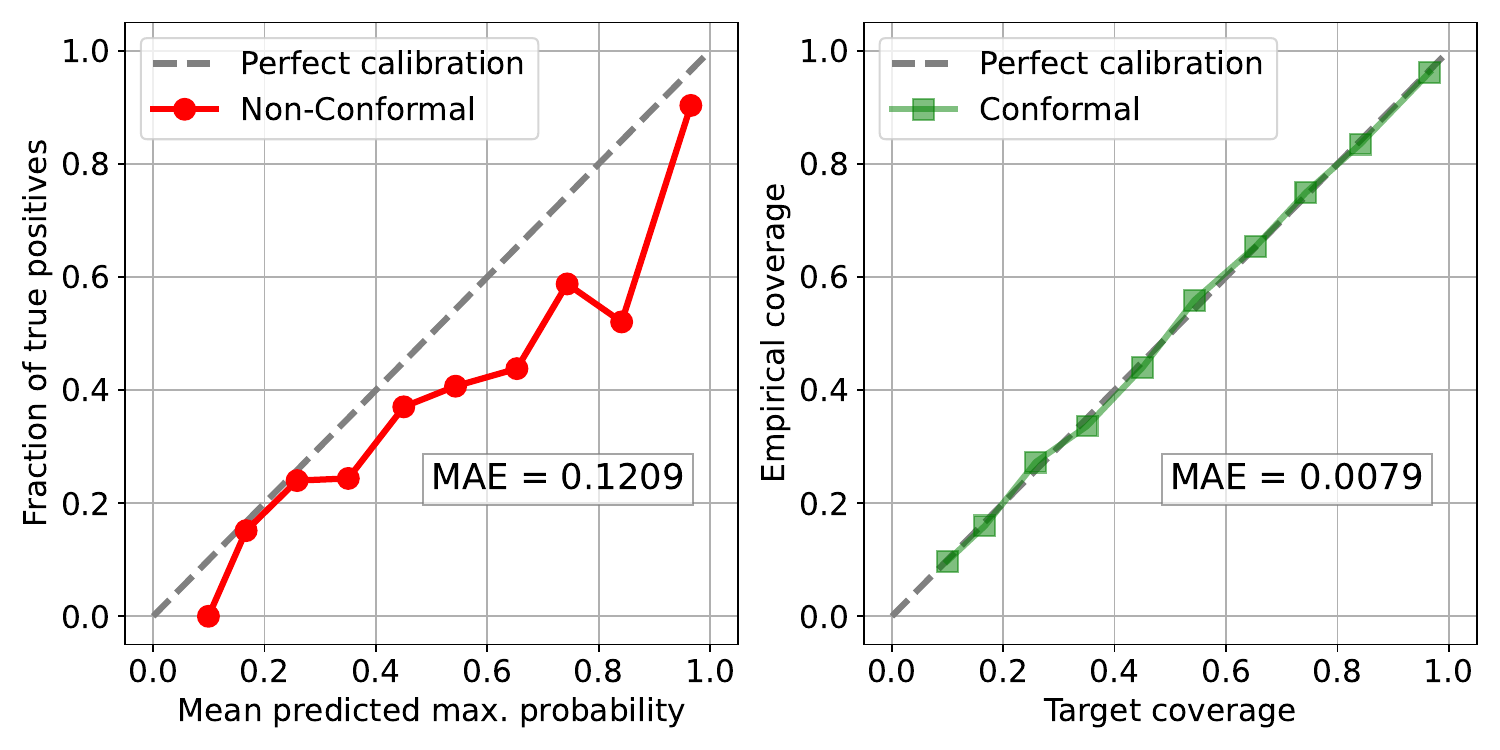}
    \caption{Conformal model is better calibrated than non-conformal model. The left plot shows the calibration curve of the pointwise classifier. For 10 equidistant bins between 0 and 1, we are plotting the ratio of true positives that fall into that bin. For a perfectly calibrated model, the true positive ratio should match the mean predicted maximum softmax probability per bin. The plot on the right displays the empirical coverage rate (how often the true label is covered by the predicted label set) against 10 target coverage rates (chosen similar to the x coordinates of the red dots on the left). These target coverages are, in fact, confidence levels for which we have recalibrated the model. As expected, the conformal model has a much lower Mean Absolute Error w.r.t. the perfect calibration line.}
    \label{fig:calib_curve}
\end{figure}

\paragraph{3. Conformal Classification Improves Calibration.} There is, however, still a benefit in working with conformal sets in the classification task. As shown in \autoref{fig:calib_curve}, the conformal model yields substantially better-calibrated confidence estimates than the original pointwise version. This is a desirable property in human-AI collaboration. In such settings, reliable probability estimates are often necessary to ensure trust and interpretability. Of course, given the comparable uncertainty ranking performance of Softmax entropies and conformal sets, one could also apply \textit{temperature scaling} \cite{temp_scaling} to the Softmax output to achieve better calibration while maintaining the same rankings as in the plain Softmax entropies.

\begin{figure}[htbp]
    \centering
    \includegraphics[width=0.7\linewidth]{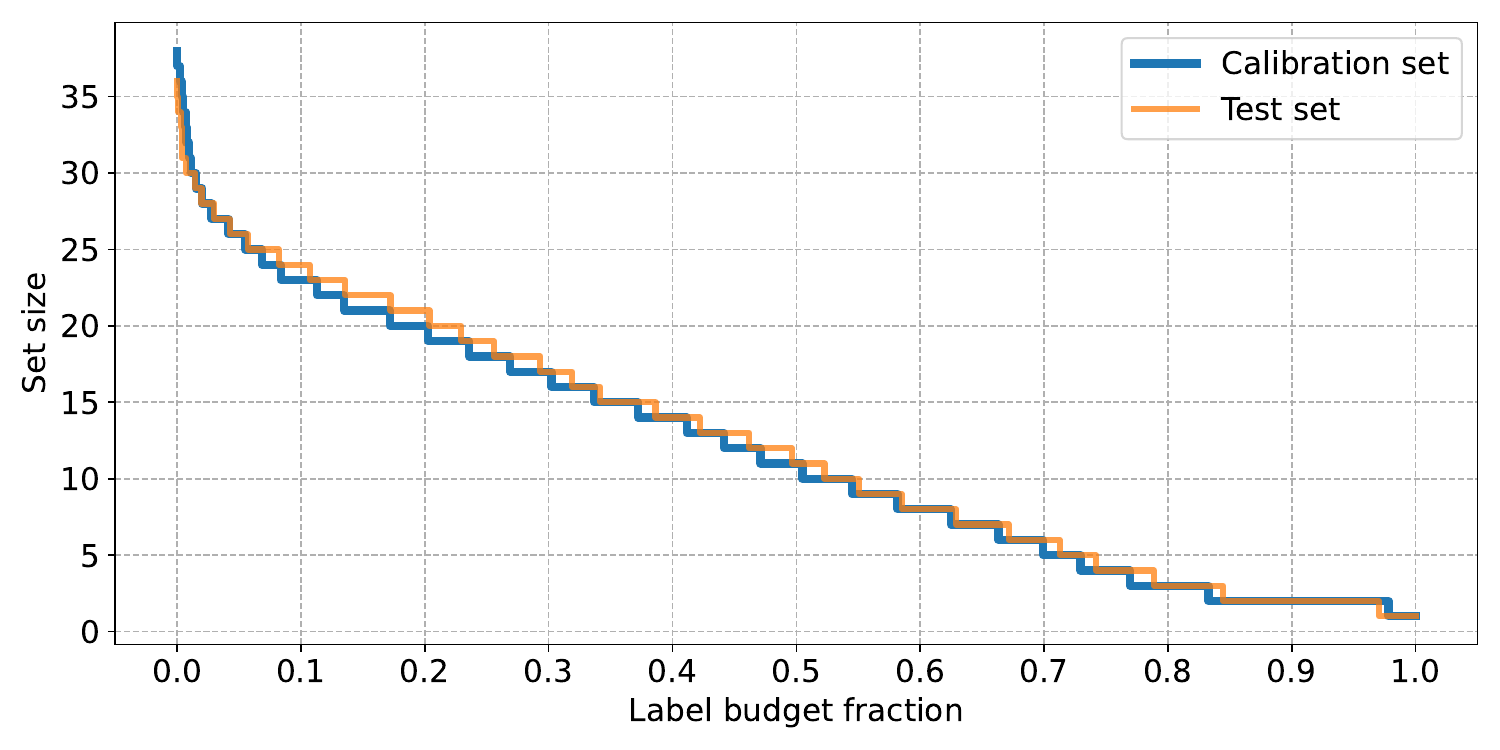}
    \caption{Cumulative distribution of predicted set sizes. In a real annotation deployment, we would look at what set size corresponds to a particular label budget and use that as a threshold for filtering difficult samples to send to expert review. For example, for a label budget of 10\%, we would filter out samples with prediction set sizes of 23 or above (blue line). The very similar distribution on the test set (orange) shows that the calibrated model generalizes well and a threshold chosen on the calibration set would be valid for new unseen data, too.
    }
    \label{fig:cumulative}
\end{figure}

\paragraph{4. Uncertainty Thresholds for Relative  Label Budgets.} Beyond the controlled experiments presented in this work, a supplementary requirement for our real-world deployment is determining uncertainty thresholds in settings where the volume of incoming data is unknown or variable. In such dynamic annotation scenarios, fixed label budgets are impractical. Instead, we propose inferring uncertainty thresholds based on the distribution of uncertainty scores - conformal interval widths or conformal set sizes — observed on the calibration set. This approach allows practitioners to translate a relative budget constraint (e.g. reviewing only the top 10\% most uncertain samples) into a concrete threshold on the uncertainty metric. For example, if the cumulative distribution of conformal set sizes (see \autoref{fig:cumulative}) indicates that the top 10\% most uncertain calibration samples have prediction sets of size 23 or greater, then this value can serve as an uncertainty threshold for expert review. This strategy offers a flexible and data-driven mechanism for adapting human-machine collaboration to evolving data scales, while respecting resource constraints.

\section{Conclusion}

In this work, we presented a conformalized version of SoilNet \cite{soilnet}, a multimodal multitask model for automated soil profile annotation. By applying conformal prediction to both regression and classification tasks, we obtained reliable well-calibrated uncertainty estimates - an essential requirement for integrating ML systems into expert-driven workflows.

To simulate real-world human–machine collaboration, we proposed an uncertainty-guided human-AI annotation framework under a constrained label budget. Our experiments demonstrate that uncertainty-aware querying, guided by conformal prediction, can improve annotation quality while minimizing expert workload. Furthermore, we showed how uncertainty thresholds can be inferred from the calibration set in accordance with restricted labeling budgets, enabling dynamic adaptation to varying volumes of incoming data.

While this study focused on depth marker regression and horizon classification, similar uncertainty-aware strategies can be applied to the other tasks SoilNet is trained to solve - morphological properties prediction. More broadly, our findings support the use of conformal prediction as a principled, practical tool for enhancing transparency and collaboration in expert-AI systems deployed in scientific and environmental domains.

\begin{acks}
We thank the Bundesanstalt für Geowissenschaften und Rohstoffe (BGR) for the financial support of this research and particularly Dr. Einar Eberhardt and Dr. Stefan Broda for their expert guidance regarding the complex geological aspects of soil horizon classification. This research was also supported by the German Research Foundation (DFG) - Project number: 528483508 - FIP 12. We also thank Sebastian Jäger for his valuable advice on issues related to conformal classification.
    
\noindent{The authors have no competing interests to declare that are relevant to the content of this article.}
\end{acks}

  

\bibliographystyle{ACM-Reference-Format}
\bibliography{references}



\end{document}